\documentclass[10pt,twocolumn,letterpaper]{article}

\usepackage{wacv}
\usepackage{times}
\usepackage{epsfig}
\usepackage{graphicx}
\usepackage{amsmath}
\usepackage{amssymb}
\usepackage{color}
\usepackage{url}
\usepackage{xcolor}
\usepackage{bm}
\usepackage{marvosym}
\usepackage{fancyhdr}

\wacvfinalcopy 

\def\wacvPaperID{256} 

\ifwacvfinal\fi
\fancypagestyle{firststyle}
{
   
   \fancyhf{}
   \fancyfoot[C]{\raggedright \footnotesize
         \copyright  2017 IEEE. Personal use of this material is permitted. Permission from IEEE must be obtained for all other uses, in any current or future media, including reprinting/republishing this material for advertising or promotional purposes, creating new
collective works, for resale or redistribution to servers or lists, or reuse of any copyrighted component of this work in other works.}
}

\begin{document}
\title{Joint Regression and Ranking for Image Enhancement}

\author{Parag Shridhar Chandakkar \\
Arizona State University\\
{\tt\small pchandak@asu.edu}
\and
Baoxin Li \\
Arizona State University\\
{\tt\small baoxin.li@asu.edu}
}

\maketitle
\ifwacvfinal\thispagestyle{empty}\fi
\thispagestyle{firststyle}
\begin{abstract}

Research on automated image enhancement has gained momentum in recent years, partially due to the need for easy-to-use tools for enhancing pictures captured by ubiquitous cameras on mobile devices. Many of the existing leading methods employ machine-learning-based techniques, by which some enhancement parameters for a given image are found by relating the image to the training images with known enhancement parameters. While knowing the structure of the parameter space can facilitate search for the optimal solution, none of the existing methods has explicitly modeled and learned that structure. This paper presents an end-to-end, novel joint regression and ranking approach to model the interaction between desired enhancement parameters and images to be processed, employing a Gaussian process (GP). GP  allows searching for ideal parameters using \textit{only} the image features. The model naturally leads to a ranking technique for comparing images in the induced feature space. Comparative evaluation using the ground-truth based on the MIT-Adobe FiveK dataset plus subjective tests on an additional data-set were used to demonstrate the effectiveness of the proposed approach.

\end{abstract}

\section{Introduction}

The corpus of images on the Web is exponentially increasing in size with close to two billion photos being added or circulated each day\footnote{\url{http://www.kpcb.com/internet-trends}}. Image sharing has become an integral part of daily life for many people, and they want their photos to look good without doing too much manual editing. Some tools for easy image enhancement have already been deployed on popular social networking platforms, such as Instagram, or mobile devices such as iPhone. However, most such tools are essentially based on some pre-defined image filters for obtaining certain visual effects. Recent research efforts on automated image enhancement employing machine learning techniques for improved functionalities such as content adaptivity and personalization. Such solutions range from learning a tone mapping between the spaces of low-quality and high-quality images\footnote{We call the images before enhancement as low-quality and those after the enhancement as high-quality in the rest of this article. We also refer to the process of enhancing a new picture as ``the testing stage".} to building a ranking relation between these two spaces \cite{berthouzoz2011framework,bychkovsky2011learning,kapoor2014collaborative,hwang2012context,kang2010personalization,kaufman2012content,chandakkar2015relative}, although we are yet to see such techniques being deployed on a popular platform.

\begin{figure*}[!t]
\centering  
\includegraphics[width=0.7\textwidth]{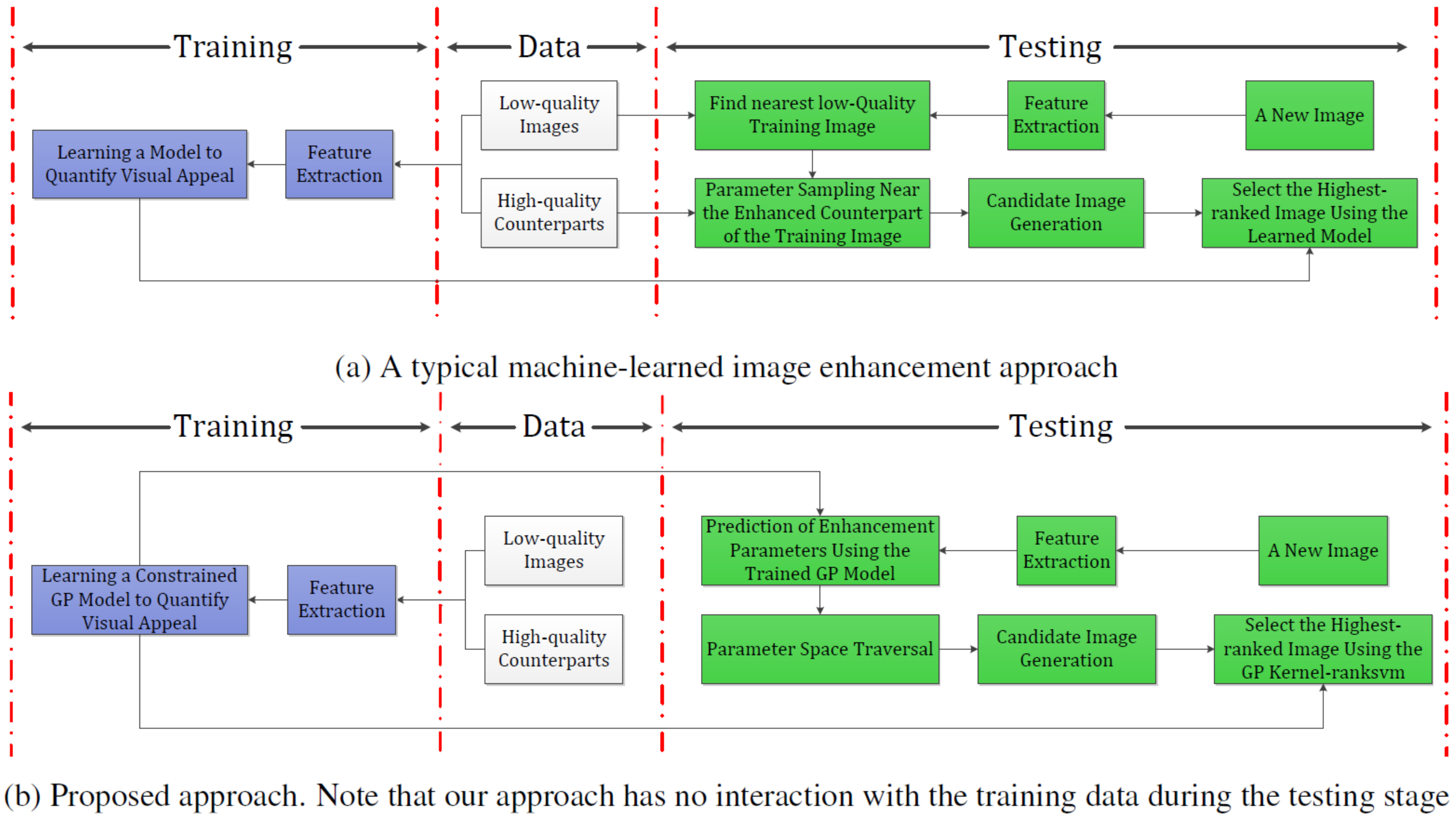}
\caption{Pipelines of image enhancement approaches.}
\label{fig:allPipelines}
\end{figure*} 

Many recent approaches follow the pipeline shown in Fig. \ref{fig:allPipelines}. During training, these approaches learn a model to assign a score for a given image quantifying its \textit{visual appeal}. To enhance a new image, a nearest training image is found. Then a dense sampling around the parameters\footnote{The brightness, saturation and contrast are referred to as ``parameters'' of an image in this article} of the high-quality counterpart of that training image gives the candidate set of enhancement parameters for the new image. A set of candidate images is then generated by applying these enhancement parameters to the new image. The next step is extracting features of all candidate images and use the learned model to select the highest-quality image.

There are two major drawbacks in the above processing flow. First, it is computationally expensive at the testing phase since a search through the entire training data is needed. The training data for such applications could be of huge size and is usually hosted on a server. Thus hundreds of thousands people querying the server per second is undesirable and uncalled-for. Second, the set of candidate parameters which would enhance the original image is found in a sub-optimal manner by doing a $k$NN search. It does not provide any structured way to search for the optimal parameters and thus it becomes necessary to search the entire training set and create a lot of candidate images, resulting a computational bottleneck for the testing phase.

In this paper, we develop a joint regression and ranking approach to address the above drawbacks. Our approach employs GP regression to predict the mean and variance of the candidate parameters \textit{only} from the feature vector of a low-quality image. We also simultaneously train a ranking model on the GP-covariance-kernel-induced feature space. To achieve this, we derive and use the dual-form of ranking SVM \cite{joachims2002optimizing} with the GP kernel integrated into it. Thus the kernel builds a relation between the image feature space and its corresponding enhancement parameter space. Along with that, the kernel learns to give more weight to the image features which are highly responsible for making an image to be of higher quality. Finally, we learn the GP kernel in such a way that all the high-quality counterparts of a low-quality image form a cluster. This allows exploration of the image parameter space in a structured manner for obtaining the optimal solution.

In the testing stage (i.e., while enhancing a new image), the model provides the expected value and variances of the enhancement parameters, drastically reducing required computation since there is no need to sift through the training set. We can generate/show some enhanced images by applying parameters that are $k$ standard deviations away from the expected values of the parameters, where $k$ is a user-defined and can be changed on-the-fly, so that user can choose from some good candidate images. The same kernel can be used to rank images if the user wants to see a single image. Through extensive experiments, we show that our approach is computationally efficient at the testing phase, and that it predicts parameters correctly for new images and that it also predicts the ranking relations between the new images and its enhanced counterparts.

\section{Related Work}

Automated image enhancement has recently been an active research area. Various solutions have been proposed for this task. We review those works which aim to improve the visual appeal of an image using automated techniques. A novel tone-operator was proposed to solve the tone reproduction problem \cite{reinhard2002photographic}. A database named MIT-Adobe FiveK of corresponding low and high-quality images was published in \cite{bychkovsky2011learning}. They also proposed algorithm to solve the problem of global tonal adjustment. The tone adjustment problem only manipulates the luminance channel. In \cite{joshi2010personal}, an approach was presented, focusing on correcting images containing faces. They built a system to align faces between a ``good'' and a ``bad'' photo and then use the good faces to correct the bad ones.

Content-aware enhancement approaches have been developed which aim to improve a specific image region. Some examples of such approaches are \cite{berthouzoz2011framework,kaufman2012content}. A drawback of these is the reliance on obtaining segmented regions that are to be enhanced, which itself may prove difficult. Pixel-level enhancement was performed by using local scene descriptors. First, images similar to the input are retrieved from the training set. Then for each pixel in the input, a set of pixels was retrieved from the training set and they were used to improve the input pixel. Finally, Gaussian random fields are used to maintain the spatial smoothness in the enhanced image. This approach does not take the global information of an image into account and hence the local adjustments may not look right when viewed globally. A deep-learning based approach was  presented in \cite{yan2014automatic}. In \cite{kang2010personalization}, users were required to enhance a small amount of images to augment the current training data.

Two closely related and recent works involve training a ranking model from low and high-quality image pairs \cite{yan2014learning,chandakkar2015relative}. In a recent state-of-art method \cite{yan2014learning}, a dataset of $1300$ corresponding image pairs was reported, where even the intermediate enhancement steps are recorded. A ranking model trained with this information can quantify the (enhancement) quality of an image. In \cite{chandakkar2015relative}, non-corresponding low and high-quality image pairs were used to train a ranking model. Both the approaches use $k$NN search at the test time to create a pool of candidate images first. After extracting features and ranking all of them, the best image is presented to the user.

Now we briefly review Gaussian process based methods which are relevant in this context. GP has been effectively used to obtain good performance for applications where complex relationships have to be learned using a small amount of data (in the order of several hundreds) \cite{urtasun2007discriminative}. In \cite{eleftheriadis2015discriminative}, it was used for view-invariant facial recognition. A GP latent variable model was used to learn a discriminative feature space using LDA prior where examples from similar classes are project nearby. In \cite{rudovic2010regression}, GP regression was used to map the non-frontal facial points to the frontal view. Then facial expression methods can be used using these projected frontal view points. Coupled GP have been used to capture dependencies between the mappings learned between non-frontal and frontal poses, which improves the facial expression recognition performance \cite{rudovic2013coupled}.

Our effort in this paper deals with enhancement considering contrast, saturation and brightness of an image. We attempt to explicitly model interactions between parameters controlling these factors and features extracted from the underlying image, employing GP.
Our approach of joint regression and ranking allows us to learn the complex mapping from the image features to the regions corresponding to desired enhancement in the parameters space, without actually generating several hundreds of enhanced candidate images. The expected value of the parameters and their standard deviations provide us with a way to systematically explore the parameters space. In the next section, we detail our proposed approach. To the best of our knowledge, this is the first attempt on incorporating GP regression and ranking for image enhancement.

\section{Proposed Approach}

Our problem is to predict the set of image parameters which would enhance a given image. Our proposed approach consists of two objectives: 1. Given a low-quality image feature, probabilistically estimate the parameters that could generate the enhanced counterpart. 2. Produce a ranking in the GP-kernel-induced feature space and thereby discover the features responsible for making an image of higher-quality.

We have pairs of low and high-quality images along with their parameters for training. The feature representation of an image will be discussed in detail in section \ref{sec:featureRepresentation}. Features of $N$ low-quality images are represented by $\bm{F}=\{\bm{f}_1,\bm{f}_2,\hdots,\bm{f}_N\}$\footnote{We represent vectors by lower-case bold letters. Matrices are represented by upper-case bold letters. Scalars are denoted by non-bold letters.}. We have $p$ high-quality versions for each low-quality image. Its features are represented by $\bm{F}^+=\{\bm{F}_1^+,\hdots,\bm{F}_N^+\}$, where $\bm{F}_i^+=\{\bm{f}_{i1}^+,\hdots,\bm{f}_{ip}^+\}$, and $\bm{f}_i,\bm{f}_{ij}^+ \in \mathbb{R}^{D \times 1} \ \forall \ i,j$. We also have $p$ sets of high-quality parameters for a low-quality image. However, for illustration, we predict parameters only for the first set. Note that we still use all the $p$ sets of high-quality images to train a ranking model. The parameters of low and high-quality images are represented by $\bm{Y}=\{\bm{y}_1,\hdots,\bm{y}_N\}$ and $\bm{Y}^+=\{\bm{y}_{1}^+,\hdots,\bm{y}_{N}^+\}$ respectively. We use three image parameters, namely, brightness, contrast and saturation and hence $\bm{y}_i,\bm{y}_i^+ \in \mathbb{R}^{3 \times 1} \ \forall \ i$. Our task is to obtain $\bm{y}_i^+$ using only $\bm{f}_i$ and $\bm{y}_i$. We predict each parameter using a separate GP. To that end, we collect the $m^{th}$ parameter of all low and high-quality images into, $\bar{\bm{y}}_m=(y_{1m},\hdots,y_{Nm})^T$ and $\bar{\bm{y}}_m^+=(y_{1m}^+,\hdots,y_{Nm}^+)^T$, respectively and train a separate GP model to predict each parameter. We now outline the proposed joint GP regression and ranking.

\subsection{GP Regression}
GPs define a prior distribution over functions which becomes a posterior over functions after observing the data. GPs assume that this distribution over functions is jointly Gaussian with a mean and a positive definite covariance kernel function. GPs provide well-calibrated, probabilistic outputs which are particularly useful and necessary in our application \cite{murphy2012machine}. If we let the prior on regression function be a GP, then it can be denoted as $GP(m(\bm{f}),\kappa(\bm{f},\bm{f}^\prime))$ where $\bm{f}$ and $\bm{f^\prime}$ are image features $\in \mathbb{R}^{D \times 1}$ as defined previously, $m(\bm{f})$ is a mean function and $\kappa(\bm{f},\bm{f}^\prime)$ is a covariance function. It is well-known that GPs are flexible enough to model an arbitrary mean. It can be shown that the posterior predictive density for a single test input is:
\begin{equation} \label{eq:prediction}
p(\bar{y}_{*m}^+ | \bm{f}_*,\bm{F},\bm{Y}) = \mathcal{N} (\bar{y}_{*m}^+ | \bm{k}_*^T \bm{K}_y^{-1} \bar{\bm{y}}_m^+,k_{**}-\bm{k}_*^T \bm{K}_y^{-1}\bm{k}_*)
\end{equation}
\noindent where $\bm{k}_*=[\kappa(\bm{f}_*,\bm{f}_1),\hdots,\kappa(\bm{f}_*,\bm{f}_N)]$, $N$ is the number of samples, $k_{**}=\kappa(\bm{f}_*,\bm{f}_*)$ and $\bm{K}_y=\bm{K}+\sigma_y^2 \bm{I}_N$. $\bm{K}$ is a kernel function between all training inputs $\bm{f}$, and $(\cdot)_*$ denotes a new data sample. The noise or uncertainty in the output is modeled by the noise variance, $\sigma_y^2$ .

It can be further shown that the log-likelihood function for a GP regression model is easily obtained by using a standard multivariate Gaussian distribution as follows,
\begin{equation} \label{eq:logLikelihood}
\begin{aligned}
\log  p(\bar{\bm{y}}^+_m | \bm{F}) &= -0.5 \left( \bar{\bm{y}}^+_m \right) ^T \bm{K}_y^{-1} \bar{\bm{y}}^+_m - 0.5 log |\bm{K}_y| - \\  &0.5Nlog(2 \pi)
\end{aligned}
\end{equation} 
\noindent We choose a standard squared exponential kernel for our application. It is as follows,
\begin{equation}
\kappa(\bm{f}_i,\bm{f}_j)=\sigma_f^2 \exp(-\frac{1}{2}(\bm{f}_i-\bm{f}_j)^T\cdot \bm{\Lambda} \cdot (\bm{f}_i-\bm{f}_j))+\sigma_y^2 \delta_{ij}
\end{equation}
\noindent Here, $\sigma_f^2$ controls the vertical scale of the regression function, $\sigma_y^2$ models uncertainty, $\bm{\Lambda}$ is a diagonal matrix with entries $\{\theta_1,\hdots,\theta_D\}$ and $\delta_{pq}$ is a Kronecker delta function, which takes the value $1$ if $p=q$, it is zero everywhere else. We call $\bm{h}=\{\sigma_f^2, \bm{\Lambda}, \sigma_y^2\}$ as hyper-parameters. It is easy to see that the prediction in Equation \ref{eq:prediction} is dependent on the kernel and in turn on the hyper-parameters: $\sigma_f, \bm{\Lambda}$ and $\sigma_y^2$. We will see later how to obtain optimal hyper-parameters. Now, we explain inclusion of ranking into our formulation.

\subsection{GP Ranking}
We build a ranking relation in the GP-kernel-induced feature space. Thus the GP kernel discovers the features responsible for making an image of higher-quality and assigns higher weight to them by adjusting the hyper-parameters. The primal form of rank SVM \cite{joachims2002optimizing} is given by:
\begin{equation}
\begin{aligned}
&\min_{w,\xi_{ij}} \frac{1}{2}\bm{w}^T\bm{w} + C \sum_{i,j} \xi_{ij}, \
\text{subject to:} \ \bm{u}_i \succ \bm{u}_j \  \forall \ (i,j)
\end{aligned}
\end{equation}
\noindent where $\bm{u}_i \succ \bm{u}_j$ indicates that $\bm{u}_i$ is ranked higher than $\bm{u}_j$.

For ranking, we observe that only building a relation between low and high-quality images does not provide good ranking accuracy on new images. The reason is that the enhanced images often possess high saturation, brightness and/or contrast. Thus the ranking model sometimes assigns a higher score to over-saturated and over-exposed images. This would not be a problem if one had intermediate information about the enhancement steps being performed \cite{yan2014learning}. Thus we deteriorate our original low-quality images by shifting the image parameters to both extremes. The amount of shifting for an image is decided by first deteriorating 20 images manually and then heuristically defining a relation between existing image parameters and the amount of shifting needed to significantly deteriorate the image. Let's call these images as \textit{poor-quality} images. We also generate $p$ poor-quality images for every low-quality image. We now have features for poor, low and high quality images, denoted by $\bm{F}^-,\bm{F}$ and $\bm{F}^+$ respectively. Primal form for our ranking model can be written as follows,
\begin{equation} \label{eq:ourFormulationPrimal}
\begin{aligned}
&\hspace{35pt}\min_{w,\xi_{ij}} \ \frac{1}{2}\bm{w}^T\bm{w} + C_1 \sum_{i,j} \xi_{ij} + C_2 \sum_{i,k} \xi_{ik}^\prime, \\
&\hspace{38pt}\text{subject to:} \ \bm{w}^T \bm{f}_{ij}^+ \geq \bm{w}^T \bm{f}_i + 1 - \xi_{ij}, \\
&\hspace{38pt}\text{subject to:} \ \bm{w}^T \bm{f}_i \geq \bm{w}^T \bm{f}_{ik}^- + 1 - \xi_{ik}^\prime, \\
&\hspace{0pt}\text{subject to:} \ \bm{w}^T \bm{f}_{ij}^+ \geq \bm{w}^T \bm{f}_{ik}^- + 1 - \xi_{ik}^{\prime\prime}, \xi_{ij},\ \xi_{ik}^\prime,\ \xi_{ik}^{\prime\prime} \geq 0 \\& \forall \ i=\{1,\hdots,N\}, \forall j=\{1,\cdots,p\}, \forall k=\{1,\cdots,p\}.
\end{aligned}
\end{equation}
\noindent Now we derive the dual form to incorporate the GP-kernel, $\kappa$. It would be cumbersome to derive the dual of Equation \ref{eq:ourFormulationPrimal} as it stands. Instead, we can define a new set of data $\bm{D}$ consisting of $\bm{f}_i-\bm{f}_{ij}^+$, $\bm{f}_{ik}^- -\bm{f}_i$ and $\bm{f}_{ik}^- - \bm{f}_{ij}^+ \ \forall \ i, j, k$. Now, $\bm{D}$ has $N^\prime=N(2p+p^2)$ elements, we can write the primal form as follows:
\begin{equation} \label{eq:compactPrimal}
\begin{aligned}
&\hspace{45pt} \min_{w,\xi_i} \ \frac{1}{2}\bm{w}^T\bm{w} + C \sum_{i} \xi_i, \\
&\text{subject to:} \ \bm{w}^T \bm{D}_i + 1 - \xi_i \leq 0, \xi_i \geq 0,  \forall i=\{1,\hdots,N^\prime\}.
\end{aligned}
\end{equation} 
\noindent We use Lagrangian multipliers to convert the above equation into an unconstrained optimization problem. 
\begin{equation} \label{eq:compactPrimalMatrixForm}
\begin{aligned}
L(\bm{w},\bm{\alpha},\bm{\beta})&=\frac{1}{2} \bm{w}^T \bm{w} + C \sum_i \xi_i \ + \\& \sum_i \alpha_i (\bm{w}^T \bm{D}_i + 1 - \xi_i) - \sum_i \beta_i \xi_i
\end{aligned}
\end{equation} 
\noindent Differentiating with respect to $\bm{w}$ and $\xi$ and equating them to zero, we get,
\begin{equation} \label{eq:diffOurDualForm}
\begin{aligned}
&\nabla_w L(\bm{w},\bm{\alpha},\bm{\beta}) = 0 \Rightarrow  \bm{w} = - \sum_i \alpha_i \bm{D}_i \\
&\nabla_\xi L(\bm{w},\bm{\alpha},\bm{\beta}) = C - \alpha_i - \beta_i = 0 \Rightarrow \alpha_i \leq C.
\end{aligned}
\end{equation}
\noindent Substituting $\bm{w}$ back into Equation \ref{eq:compactPrimal} and doing some algebraic manipulation, we get a dual maximization problem as,
\begin{equation}
\max_{\bm{\alpha}} \sum_i \alpha_i - \frac{1}{2}  \sum_i \sum_j \alpha_i \alpha_j \bm{D}_i^T \bm{D}_j, \
\text{subj. to:} \ 0 \leq \alpha_i \leq C.
\end{equation}
The inner product in the above equation can be replaced with GP kernel by employing the kernel trick. Thus the final optimization problem to get $\bm{\alpha}$ becomes,
\begin{equation} \label{eq:ourDualMatrixForm}
\max_{\bm{\alpha}} \ \bm{1}^T \bm{\alpha} - \frac{1}{2} \bm{\alpha}^T \bm{K}_y \bm{\alpha}.
\end{equation}
\noindent Here, $\bm{1}$ is a column vector of ones. The length of both $\bm{\alpha}$ and $\bm{1}$ is $N(2p+p^2)$. The dimensions of $\bm{K}_y$ are $N(2p+p^2) \times N(2p+p^2)$. The $(i,j)^{th}$ element of $\bm{K}_y$ is $\kappa(\bm{D}_i,\bm{D}_j)$.

\subsection{Clustering high-quality images together}
We turn our attention to the third constraint. Given a low-quality image: 1. it forces all its high-quality counterparts to form a cluster and 2. it tries to maximize the distance between poor-quality and high-quality images in the GP-kernel-induced feature space. The intuition behind this is as follows. Ultimately, given a new image, we would not only like to predict the parameters for its enhanced counterpart, but we also wish to traverse the parameter space and explore more of such enhancement parameters. The traversing of the parameter space is, in our opinion, essential since the choices of people vary by a great amount and no model would do justice with just one set of predicted parameters. Note that this constraint tries to minimize distance between $\bm{f}_i$ and $\bm{f}_{ij}^+ \ \forall j$, so by definition of GP, the distance between the corresponding output parameters, $\bm{y}_{ij}^+ \ \forall j$, will be reduced, which in turn achieves the aforementioned effective traversal. The second part of the constraint tries to push the predicted parameters away from the parameters of the poor-quality images. The details of traversing the parameter space after getting the GP predictions are discussed later. The constraint can be formulated as follow.

\begin{equation} \label{eq:clusterConstraint}
\min_h \left( \sum_i ||\bm{K}_y^{\bm{F}_i^+}||_F^2 - ||\bm{K}_y^{\bm{F}_i^+,\bm{F}_i^-}||_F^2 \right),
\end{equation}

\noindent where $||\cdot||_F^2$ indicates squared Frobenius norm. The term $\bm{K}_y^{\bm{F}_i^+,\bm{F}_i^-}$ is a $p \times p$ matrix defined as follows,
\begin{equation}
\bm{K}_y^{\bm{F}_i^+,\bm{F}_i^-} = 
 \begin{bmatrix}
  \kappa(\bm{f}_{i1}^+,\bm{f}_{i1}^-) & \cdots & \kappa(\bm{f}_{i1}^+,\bm{f}_{ip}^-) \\
  \kappa(\bm{f}_{i2}^+,\bm{f}_{i1}^-) & \cdots & \kappa(\bm{f}_{i2}^+,\bm{f}_{ip}^-) \\
  \vdots  & \ddots & \vdots  \\
  \kappa(\bm{f}_{ip}^+,\bm{f}_{i1}^-) & \cdots & \kappa(\bm{f}_{ip}^+,\bm{f}_{ip}^-) \\
 \end{bmatrix}
\end{equation}

\noindent The term $\bm{K}_y^{\bm{F}_i^+}$ is equal to $\bm{K}_y^{\bm{F}_i^+,\bm{F}_i^+}$.

We combine Equations \ref{eq:logLikelihood}, \ref{eq:ourDualMatrixForm} and \ref{eq:clusterConstraint} to form our objective function. It is as follows,

\begin{equation} \label{eq:finalObjectiveFunction}
\begin{aligned}
\min_h \ Z = &\frac{1}{2} \left( \bar{\bm{y}}^+_m \right)^T  \bm{K}_y^{-1} \bar{\bm{y}}^+_m + \frac{1}{2} \log |\bm{K}_y| -\bm{1}^T \bm{\alpha} + \\& \frac{1}{2} \bm{\alpha}^T \bm{K}_y \bm{\alpha} \ + \sum_i \left( ||\bm{K}_y^{\bm{F}_i^+}||_F^2 - ||\bm{K}_y^{\bm{F}_i^+,\bm{F}_i^-}||_F^2 \right)
\end{aligned}
\end{equation}

\noindent Note that we have removed the constant term. We now focus on how to solve Equation \ref{eq:ourDualMatrixForm} and \ref{eq:finalObjectiveFunction} to get $\bm{\alpha}$ and $\bm{h}$.

\subsection{Optimization}

Our optimization problem is separable in $\bm{\alpha}$ and $\bm{h}$. First we optimize $\bm{\alpha}$, which can be done by using a standard rank-SVM solver. It could also be solved by using quadratic programming, however, that would be memory inefficient. In particular, we use a rank-SVM implementation which uses the LASVM algorithm proposed in \cite{bordes2005fast}. LASVM employs active example selection to significantly reduce the accuracy after just one pass over the training examples. 

After optimizing $\bm{\alpha}$, we turn our attention to Equation \ref{eq:finalObjectiveFunction}. We find its local minimizer, $\bm{h}^*$, by using scaled conjugate gradient descent (SCG) algorithm. SCG is chosen due to its ability to handle tens of thousands of variables. SCG has also been widely used in previous approaches involving GPs \cite{rasmussen2006gaussian,eleftheriadis2015discriminative,rudovic2013coupled}. We use chain rule to compute $\frac{\partial Z}{\partial \bm{h}}$ by evaluating first $\frac{\partial Z}{\partial \bm{K}_y}$ and then $\frac{{\partial \bm{K}_y}}{\partial \bm{h}}$. The matrix calculus identities from \cite{petersen2008matrix} are used while computing the following expressions.

\begin{equation} \label{eq:allDerivatives}
\begin{aligned}
\frac{\partial Z}{\partial \bm{K}_y} = -\frac{1}{2} \bm{K}_y^{-1} &\bm{y}_m^+ \left(\bm{y}_m^+\right)^T \bm{K}_y^{-1} + \frac{1}{2} \bm{K}_y^{-1} + \frac{1}{2} \bm{\alpha} \bm{\alpha}^T  + \\& 2 \sum_i \left( \bm{K}_y^{\bm{F}_i^+} - \bm{K}_y^{\bm{F}_i^+,\bm{F}_i^-}\right), \\
\left[\frac{\partial \bm{K}_y}{\partial \theta_q}\right]_{ij} = -\frac{1}{2}   \sigma_f^2 &  \exp \left(-\frac{1}{2}(\bm{f}_i-\bm{f}_j)^T \Lambda  (\bm{f}_i-\bm{f}_j) \right) \cdot \\& (\bm{f}_i^{(q)}-\bm{f}_j^{(q)})^2, \\
\frac{\partial \bm{K}_y}{\partial \sigma_f^2} = \sigma_f^2 \exp &(-\frac{1}{2} (\bm{f}_i-\bm{f}_j)^T\cdot \Lambda \cdot (\bm{f}_i-\bm{f}_j)),\\
&\left[\frac{\partial \bm{K}_y}{\partial \sigma_y^2}\right]_{ij} = \delta_{ij}, \\
&\hspace{-70pt}\frac{\partial Z}{\partial \theta_q} = \text{tr} \left[ \left( \frac{\partial Z}{\partial \bm{K}_y} \right)^T \left(   \frac{\partial \bm{K}_y}{\partial \theta_q} \right) \right] \ \ \forall q \in \{1,\hdots,D\},
\end{aligned}
\end{equation}

\noindent where $\text{tr}$ denotes matrix trace. Similarly, $\frac{\partial Z}{\partial \sigma_f^2}$ and $\frac{\partial Z}{\partial \sigma_y^2}$ are computed to construct $\frac{\partial Z}{\partial \bm{h}} \in \mathbb{R}^{D+2}$. This derivative can now be used to obtain the optimal set of hyper-parameters, $\bm{h}$. In practice, all the matrix inverses are implemented using Cholesky decomposition. We alternately optimize for $\bm{\alpha}$ and $\bm{h}$ till Equation \ref{eq:finalObjectiveFunction} converges or the maximum cycles are reached. We set the convergence criterion to be $10^{-3}$ and the maximum cycles to 20. 


\subsection{Testing}

Once we get the optimal $\bm{\alpha}$ and $\bm{h}$, we can predict the parameters, $\{\bar{y}_{*1}^+,\bar{y}_{*2}^+,\bar{y}_{*3}^+\}$, for the enhanced counterpart by using three trained GP models in Equation \ref{eq:prediction}. Let us call the mean and variances of the predicted parameters as $\bm{m}=\{m_1,m_2,m_3\}$ and $\bm{s}=\{s_1,s_2,s_3\}$ respectively. With their availability, we now explain our parameter space traversal.

As mentioned before, people's choices vary a lot in such applications. Thus, it is essential to explore the parameter space to generate additional enhancement parameters. The first advantage of our approach is that we can generate such parameters without referring to the training set. Since we explore the parameter space in a structured manner (with a certain mean and variance), we can afford to generate only $32$ parameters per image instead of hundreds as done in conventional $k$NN-based heuristic methods.

The First step in parameter space traversal is to determine lower and upper bounds. Those can be decided heuristically. For example, we decrease the saturation, brightness and contrast at most by an amount of $\{15\%,15\%,5\%\}$ and increase it at most by $\{35\%,35\%,20\%\}$ of the original image parameter values. We observed that these limits are not absolutely critical to the quality since the generated images will be ranked later using the learned $\bm{\alpha}$ and the images with extreme parameter settings will usually be filtered out. 

Now, we change (increase and decrease) the mean value of the parameters by $\bm{\mu}\bm{s}$ till it reaches the pre-specified thresholds. Intuitively, we think that $\bm{s}$ gives us the direction of our stride in the parameter space and $\bm{\mu}$ gives us the length of that stride. The value of $\bm{\mu}$ is determined by the number of enhanced counterparts the user wants to generate for each low-quality image. We set that value to be $30$. This value could be decreased if the user is on a mobile device with a smaller screen and similarly increased when operating on a desktop. These settings can be changed on-the-fly.

\subsection{Image feature representation} \label{sec:featureRepresentation}

We extract $432$-D color histogram with 12 bins for hue, 6 bins each for saturation and value, which acts as a global feature. We then divide the image into a $12 \times 12$ grid. For each grid, we calculate its saturation, value by taking the mean values of those image blocks in the HSV color space. We also calculate RMS contrast on that grid. These act as localized features of $144$-D each. We finally append the image parameters, which are average saturation, value and RMS contrast. Appending the image parameters allows GP to express the parameters of the enhanced counterparts as a function of both, the low-quality parameters and its feature vector. We finally get a $867$-D $(=432+3 \times 144 + 3)$ representation for every image.

\subsection{Implementation Details and Efficiency}

GPs are known to be computationally-intensive. They take about $O(N^3)$ time for training, where $N$ are the number of training examples. The matrix inversion of an $N \times N$ matrix and the computation of the derivative of the kernel are the bottlenecks in the GP training procedure. We train a GP model using about $1200$ low-quality images and six counterparts per image in about $18$ hours on an Intel Xeon \MVAt 2.4 GHz $\times$ 16. The computational efficiency can be improved by using GP regression techniques proposed for large data \cite{hensman2013gaussian,ambikasaran2014fast} or using efficient data-structures such as kd-trees \cite{shen2006fast}. During testing, our approach is extremely fast. We tested it on two systems, Intel Xeon and a modern desktop system with Intel i7 \MVAt 3.7GHz. It can predict all the three parameters for $3150$ and $1287$ images per second using Intel Xeon and i7 systems respectively. A built-in $k$NN-search function processes only $224$ images per second when asked to find one nearest neighbor in $5000$ image data-set on the Intel Xeon system. All the implementations are done in MATLAB. Since our approach need not query the training database, it could be portable and potentially allow for enhancements being performed on mobile devices.

\section{Data-sets and Experimental Setup}

In this section, we present describe the data and experimental setup. Results of these experiments are presented in the Section \ref{sec:results}.
We perform four kinds of experiments. The first experiment provides a weak quantitative measure of the accuracy of our approach. We use the MIT-Adobe FiveK \cite{bychkovsky2011learning} data-set for this experiment. This data-set has $5000$ low-quality images with $5$ expert-enhanced counterparts for each image. This is the largest such data-set available. We use $1200$ images and 6 counterparts (three each for poor and high-quality) per low-quality image to train our GP models. We use $1500$ and $800$ images for validation and testing respectively. We predict the parameters (i.e. brightness, contrast and saturation) for the first enhanced counterpart of all the images in the test set. Then we calculate the root mean square error (RMSE) and a more stringent criterion - Pearson's correlation - between the ground-truth parameters computed from expert-enhanced image and our predicted parameters. We compare our quantitative results against twin Gaussian processes (TGP) \cite{twinGP}. TGP is a strutured prediction method which considers correlation between both input and output to produce predictions. Though a low RMSE between ground truth and predicted parameters does not guarantee that the enhancement will be visually appealing (unless the RMSE tends to zero), it gives us a confirmation that the prediction is lying near the ground-truth in the parameter space. Also, this experiment validates the effectiveness of the GP regressor.

Second experiment is a qualitative measure of the image quality produced by the proposed and the competing algorithms - $k$NN, Picasa and \cite{yan2014learning}. The metric of $L2$ error in the L*ab space was adopted in \cite{yan2014learning}. We believe that it is a poor indicator of the enhancement quality and instead opt for Visual Information Fidelity (VIF) metric \cite{sheikh2006image}. This metric has the ability to predict whether the visual quality of the other image has been enhanced in comparison with the reference image by producing a value greater than one. This is unlike other quality metrics such as SSIM \cite{wang2004image}, FSIM \cite{zhang2011fsim}, VSI \cite{zhang2014vsi} etc. We use the publicly available implementation of VIF\footnote{available at \url{live.ece.utexas.edu/research/quality/}}. We calculate the VIF between the proposed enhancement (reference image) and the enhancement by 1. $k$NN 2. Picasa and 3. the approach of \cite{yan2014learning} (other image). Thus VIF $< 1$ implies that the proposed enhancement is better than the one produced by the competing algorithm and vice-versa. This comparison is done for 60 pairs where 15 images each are enhanced using Picasa and \cite{yan2014learning}, whereas the remaining 30 images are enhanced using $k$NN approach.

Third experiment is aimed towards evaluating the effectiveness of GP ranking. For each image we generate only 32 enhanced versions. Our GP ranker selects the highest ranked image out of those 32 and presents it to the user. The highest ranked image is supposed to have the best quality. We compute the VIF metric between the best image selected by the ranker (reference image) and the other 31 images (other images). Ideally, for all these 31 images, we should get values less than one indicating that GP ranker has indeed selected the best image.
 
We also carry out a subjective evaluation test to assess if people prefer the enhanced counterparts generated by our approach. We compare our approach against three other methods. First one is the $k$NN-based approach. Given a low-quality image, we search for the nearest non-duplicate image from the $5000$ images of MIT-Adobe dataset. The parameters of the expert-enhanced counterparts of the nearest image are applied to the given low-quality image. In this manner, we generate $5$ enhanced counterparts per low-quality image. Note that, $k$NN utilizes all other $4999$ images whereas we only use the model trained on $1200$ images for prediction. Then we compare against Picasa's one-touch-enhance tool. The third approach is from \cite{yan2014learning}, which also is a learning-to-rank based image enhancement approach that uses the pipeline shown at the top in Fig. \ref{fig:allPipelines}.

We use $60$ images for the subjective test which was performed by 15 people. Thirty images are selected from our testing set of the MIT-Adobe data-set. The rest of the images are from the data-set used in the paper \cite{yan2014learning}. Since we only have access to their testing set, we use that data-set solely for subjective test purposes. It contains $124$ images out of which we randomly select $30$ images. We enhance all the $60$ images using our approach. The comparison against other methods is done as follows.

The first $30$ images from the MIT-Adobe data is split into two halves. The first half is enhanced using the $k$NN approach and the second half is enhanced using Picasa. The remaining 30 images from \cite{yan2014learning} are split into two halves. The first half is enhanced using the $k$NN approach and for the second half, we directly use the high-resolution results of the test data-set of \cite{yan2014learning}. Thus each person compares 60 image pairs. One of the image in that pair has been enhanced using our approach and the other image has been enhanced using either $k$NN approach, Picasa or the approach of \cite{yan2014learning}. The subject has to choose the image which he/she finds ``visually-appealing''. If the subject feels that both images have almost the same visual appeal, a third option of preferring neither image is provided. The order in which the images appear in front of a subject is always randomized. The pairing order is also randomized. The subjects do the evaluation test in standard lighting conditions and at a comfortable and constant distance from the screen.

\section{Results} \label{sec:results}

We present results of our quantitative analysis first. We train three GP models to predict saturation, brightness and contrast for $800$ images from the test set of MIT-Adobe data-set. When compared with the parameters of expert-enhanced counterparts, we achieve RMSEs of $0.0057, 0.0022, 0.0037$ and correlations of $0.5359, 0.5553, 0.8023$ respectively, for the above three parameters. TGP gets an average RMSE of $0.0022$ but it suffers while producing an average correlation of only $0.3326$. We can see that it is relatively easier for a GP to relate the contrast to the image quality, which is intuitive since contrast variation changes the image drastically and it also makes the image look vibrant or dull. This in turn contributes most to the visual appeal of an image.

\begin{figure}[!t] 
\centering
\includegraphics[width=0.4\textwidth]{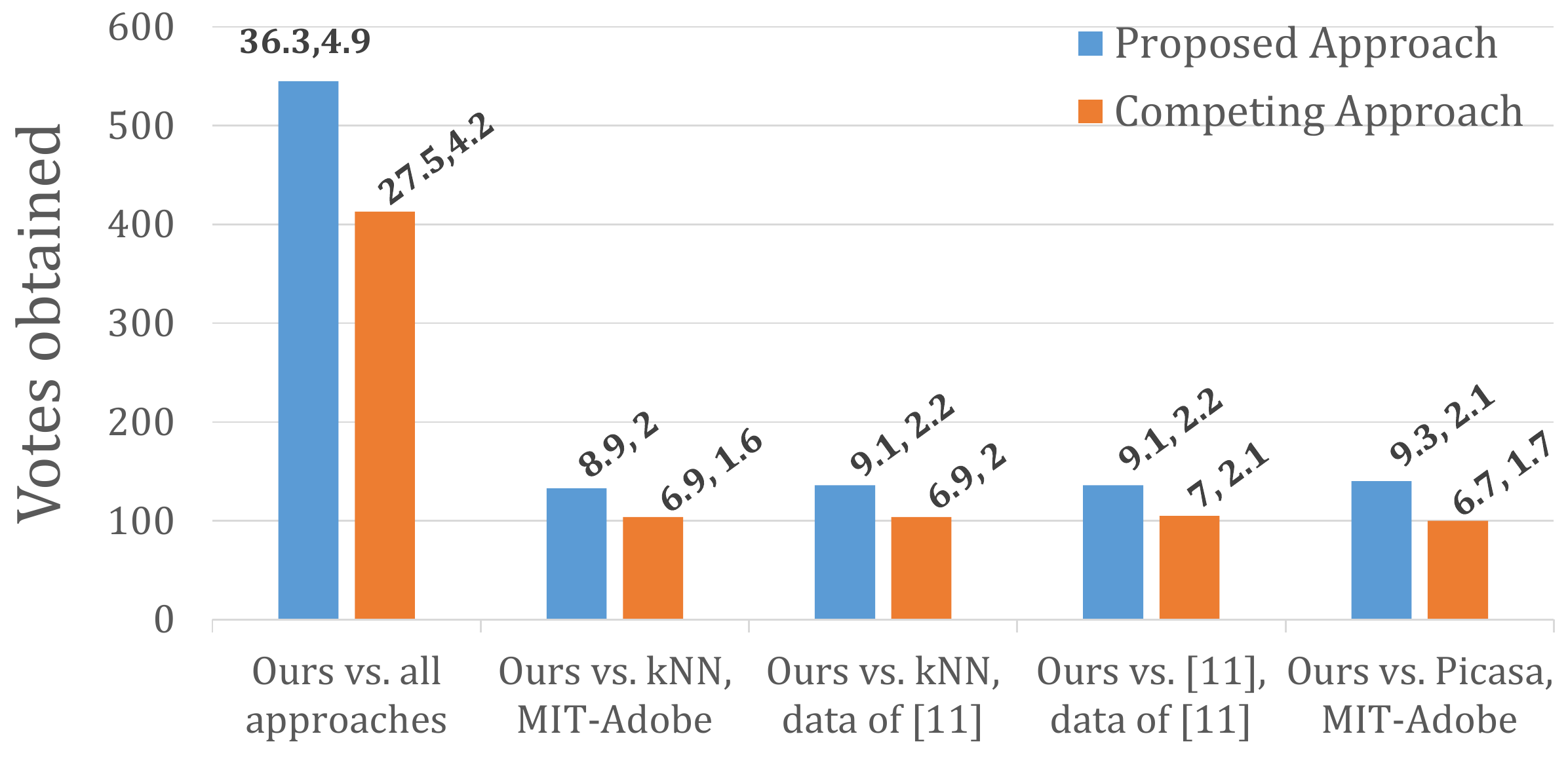}
\caption{Subjective evaluation test metrics.}  \label{fig:results}
\end{figure}

\begin{figure*}[!t] 
\centering
\includegraphics[width=0.6\textwidth]{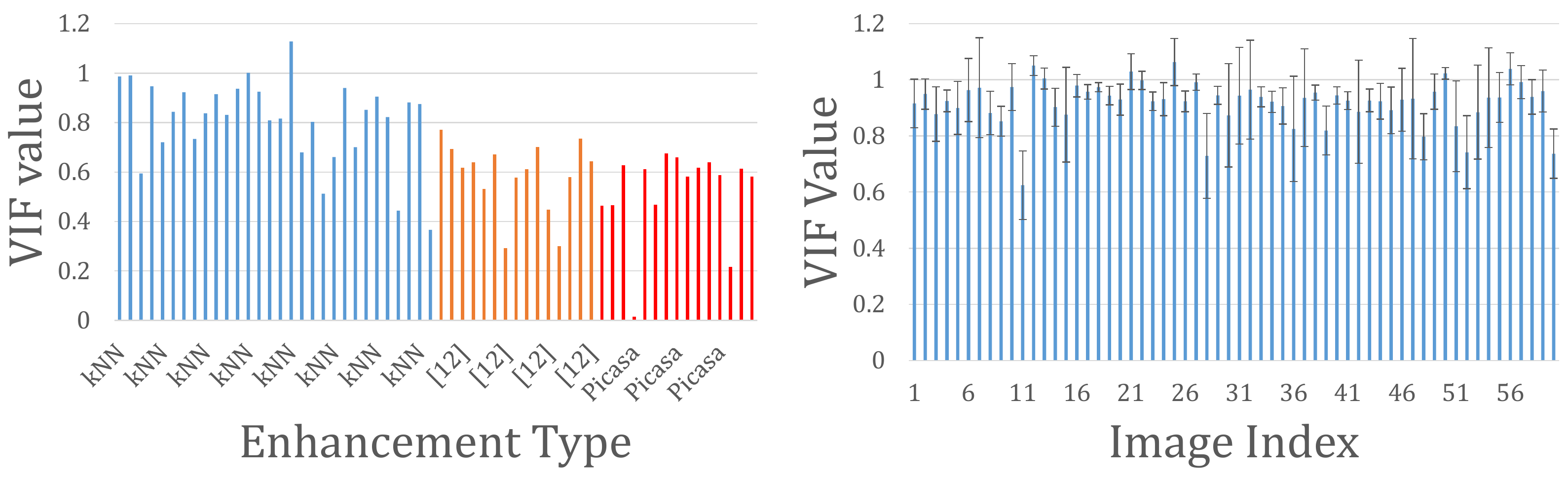}
\caption{Left plot shows VIF values comparing proposed enhancement and enhancements produced by competing algorithms. The right plot shows the mean and standard deviation of the VIF values between the best enhancements and 31 other enhancements ``rejected'' by GP ranker. VIF values $< 1$ are desirable in both the cases.}  \label{fig:barPlotExps}
\end{figure*}

\begin{figure*}[!t] \centering
\includegraphics[width=0.78\textwidth]{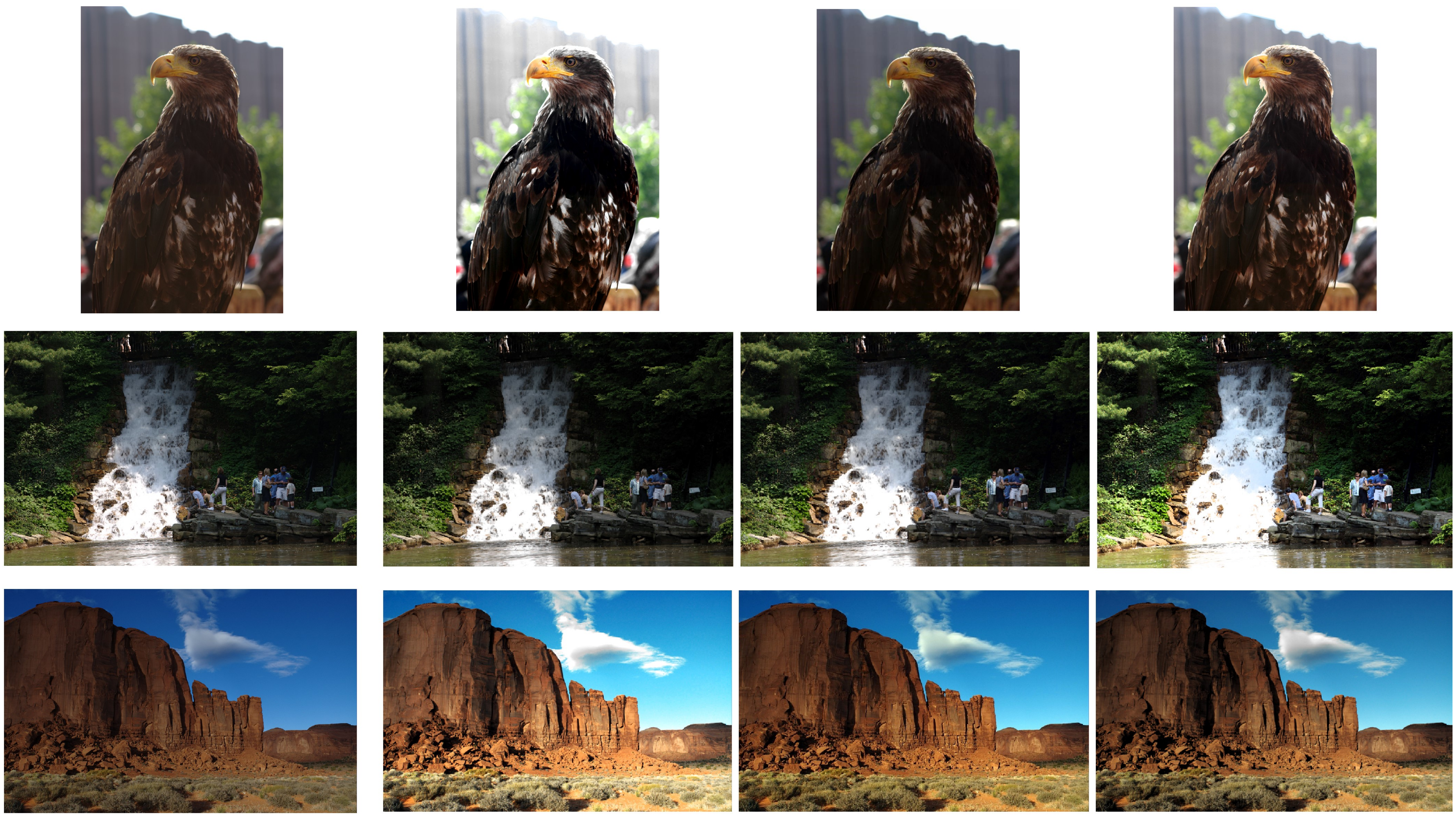}
\caption[]{The left column always contains an original low-quality image. \textbf{Row 1 and 3}: Columns 2-4 contain images enhanced by $k$NN, Picasa and GP respectively. \textbf{Row 2}: The right three columns contain enhanced versions generated by GP. Please read text for details.}
\label{fig:visualResults}
\end{figure*}

The left bar chart in Fig. \ref{fig:barPlotExps} shows the results of second experiment. VIF between our enhancement and competing enhancements produces values which are, in most cases, less than one. Thus according to VIF metric, our approach produces better enhancements than Picasa, $k$NN-based heuristics and \cite{yan2014learning}. For third experiment, we get 32 VIF values for each image, which correspond to 32 enhanced versions generated by our approach. The GP ranker selects one, as mentioned earlier. We compute the average VIF value and its standard deviation over 31 other images. This process is repeated for all the 60 images and the VIF values are shown in the right bar chart of Fig. \ref{fig:barPlotExps}.

We now analyze the results of our subjective tests. We provide the following five metrics about our subjective test in Fig. \ref{fig:results}. 1. we count votes gathered by our approach and by all other competing approaches bundled into one. This is a coarse measure of how much preference people have towards enhancements generated by our approach. 2. We count votes gathered by our approach and by $k$NN approach on the MIT-Adobe data-set. 3. comparison of votes gathered by our approach and by the $k$NN approach on the data-set of \cite{yan2014learning}. 4. comparing our approach against the results of \cite{yan2014learning} on their data. 5. Lastly, we compare our approach versus Picasa on the MIT-Adobe data. Fig. \ref{fig:results} shows all these metrics. On top of each bar, we indicate the mean and standard deviation for that particular approach and metric. For example, the second set of bars denote that for the MIT-Adobe data, our approach gathered $133$ votes against $104$ votes gathered by the $k$NN approach. The average number of votes obtained per user for our and the $k$NN approach were $8.9$ and $6.9$ with the standard deviations of $2$ and $1.6$, respectively. Fig. \ref{fig:results} shows that people consistently prefer our approach over other state-of-art approaches.

Fig. \ref{fig:visualResults} shows some of the results obtained by ours and the approach of \cite{yan2014learning}, $k$NN and Picasa's auto-enhance tool. The first and the third row illustrate that the $k$NN approach is not always effective and sometimes may give over(under)-exposed results due to its dependence on the nearest training image parameters. The second row shows three representative versions generated by GP. We can see that the image in the fourth column is over-exposed. However, our ranking model successfully filters out that image and selects the one in the third column. In general, we observed that $k$NN can only get comparable results to Picasa and our approach if it finds a good match in the training set. Thus $k$NN is unlikely to scale to large-scale enhancement tasks.

\section{Conclusion}

We presented a novel approach to image enhancement using joint regression and ranking by employing GPs. We train our GP models on the pairs formed from poor, low and high-quality images. The learned GP models predict the desired parameters for a low-quality image from its features, which may produce its enhanced counterparts. We also described a strategy to traverse the parameter space without referring to the training images, which makes our approach efficient during testing. The GP prediction is defined by the covariance kernel, on which we impose two constraints. The first one enables the kernel to learn the feature dimensions responsible for making an image of higher-quality. The other constraint clusters all the enhancement parameters corresponding to a low-quality image, thereby allowing for effective parameter traversal. We perform quantitative and subjective evaluation experiments on two-data sets to assess the effectiveness of our approach, first one being the MIT-Adobe data \cite{bychkovsky2011learning} and the another one proposed in \cite{yan2014learning}. Quantitative experiments show that our predictions produce a low RMSE when compared with the ground-truth parameters of the MIT-Adobe data. The results show that people consistently prefer the enhancements produced by the proposed approach over the other state-of-art approaches.

\textbf{Acknowledgement:} The work was supported in part by ONR grant N00014-15-1-2344 and ARO grant W911NF1410371, and a gift from Nokia. Any opinions expressed in this material are those of the authors and do not necessarily reflect the views of the sponsors.

{\small
\bibliographystyle{ieee}
\bibliography{egbib}
}

\end{document}